# COARSE-GRAINED DECOMPOSITION AND FINE-GRAINED INTERACTION FOR MULTI-HOP QUESTION ANSWERING

Xing-Cao[1,2], Yun-Liu[1,2]

[1] School of Electronic and Information, Beijing Jiaotong University,

Beijing 100044, China

(18111006, liuyun) @bjtu.edu.cn

[2] Key Laboratory of Communication and Information Systems, Beijing Municipal Commission of Education,

Beijing 100044, China

**Abstract**

Recent advances regarding question answering and reading comprehension have resulted in models that surpass human performance when the answer is contained in a single, continuous passage of text, requiring only single-hop reasoning. However, in actual scenarios, lots of complex queries require multi-hop reasoning. The key to the Question Answering task is semantic feature interaction between documents and questions, which is widely processed by Bi-directional Attention Flow (Bi-DAF), but Bi-DAF generally captures only the surface semantics of words in complex questions, and fails to capture implied semantic feature of intermediate answers. As a result, Bi-DAF partially ignores part of the contexts related to the question and cannot extract the most important parts of multiple documents. In this paper we propose a new model architecture for multi-hop question answering, by applying two completion strategies: (1) **C**oarse-**G**rain complex question **De**composition (CGDe) strategy are introduced to decompose complex question into simple ones under the condition of without any additional annotations (2) **F**ine-**G**rained **In**teraction (FGIn) strategy are introduced to better represent each word in the document and extract more comprehensive and accurate sentences related to the inference path. The above two strategies are combined and tested on the SQuAD and HotpotQA datasets, and the experimental results show that our method outperforms state-of-the-art baselines.

## 1 Introduction

One of the long-standing goals of natural language processing (NLP) is to build systems capable of reasoning about the information present in text. Tasks requiring reasoning include question answering (QA)[1,2], machine reading comprehension[3,4] (MRC), dialogue systems[5,6], and sentiment analysis[7]. Reading comprehension and question answering, which aim to answer questions about a document, have recently become a major focus of NLP research. Several different QA datasets have been proposed, such as the Stanford Question Answering Dataset (SQuAD) [8,9], NarrativeQA [10] and CoQA[11], and this kind of reasoning is termed single-hop reasoning, since it requires reasoning over a single piece of evidence. Recent advances regarding QA and MRC have surpassed human performance on some single-hop datasets, but those datasets have gaps from real-world scenarioes.

A more challenging and real-world application task, called multi-hop reasoning [12], requires combining evidence from multiple sources, which means that evidence can be spread across multiple paragraphs. In the process of reasoning, a subset of these paragraphs may be read first to extract the useful information from the other paragraphs, which might otherwise be understood as not completely relevant to the question. There exist several different datasets that require multi-hop reasoning in multiple documents, such as HotpotQA [13] and WikihopQA [14].

| |
|---|
| **Q** The rapper whose debut album was titled "Thug Misses" has sold over how many records worldwide? |
| **P1** '**Thug Misses is the debut album by American rapper Khia.**', ' The album was originally released in the United States on October 30, 2001… <br> **P2** 'Khia Shamone Finch (born Khia Shamone Chambers, November 8, 1970), …' **To date Khia has collectively sold over 2 million records worldwide**.' |
| **Q1** Who is the rapper whose debut album was titled 'Thug Misses'? <br> **Q2** How many records has that rapper sold worldwide? |

Table 1: An example of a multi-hop question from HotpotQA. The first cell shows given complex question; at the bottom of the cell are two simple questions that have been solved. The second cell contains the supporting sentences (boldface part) needed to answer the question (support facts); the highlighted part is the final answer.

As shown in Table 1, the model with strong interpretability has the ability to find supporting facts (the boldface part in P1 and P2) of the answer while the answer itself is identified. In a sense, the supporting facts predicted task is also a demonstration of the reasoning process.

Multi-hop QA faces two challenges. The first is the difficulty of reasoning due to the complexity of the query. For this challenge, some embedding-based models used to decompose query or generate query (Min et al., 2018[15]; Qi et al., 2019[16]) have been proposed, it is easier to find answers by breaking down complex questions into simple ones; for example, the question in Table 1 can be decomposed into two subquestions "Who is the rapper whose debut album was titled 'Thug Misses'?" and "How many records has that rapper sold worldwide?", but most existing work decomposes questions using a combination of rule-based algorithms, hand-crafted heuristics, and learning from supervised decompositions, each of which require significant human effort.

The second challenge is the interpretability of the model. Jiang et al. [17] pointed-out that models can directly locate the answer by word-matching the question with a sentence in the context, in which examples contain reasoning shortcuts. Then, finding all the supporting facts (inference paths) is equally important for multi-hop inference tasks.

To solve these two problems, the decomposition of complex queries and fine-grained feature interactions between documents and query are considered important for models based on semantic features. Inspired by the existing model proposed by Min et al. [15], we propose two novel completion strategies called the **C**oarse-**G**rain **De**composition (CGDe) strategy and **F**ine-**G**rained **In**teraction (FGIn) strategy. The CGDe is used to achieve better predictive capacity and explainability for question decomposition without any additional annotations, and the FGIn is used to better represent each word in the document which helps the model extract more comprehensive and accurate sentences needed to answer the question.

Different from previous works, we aims to use lightweight models instead of using off-the-shelf grammatical tools to perform grammatical processing such as named entity recognition for the construction of graph networks. Because any model that removes documents which are not related to queries will definitely improve the model effect, we are not committed to filtering irrelevant documents in advance, but seek to control the amount of passage information in the hidden representations directly.

To summarize, the key contributions are three-fold: (1) The coarse-grained complex question decomposition strategy decomposes the complex queries into simple queries without any additional annotations. (2) The fine-grained interaction strategy is used to extract more comprehensive and accurate sentences related to the inference path (3) Our model is validated on multi-hop QA and single-hop QA datasets, and the experimental results show that the model can preserve or even surpass the original system in the objective evaluations, in addition to enhancing the interpretability of the reasoning process.

## 2 Related Work

**Single-hop Question Answering**

Most MRC datasets require single-hop reasoning only, which means that the evidence necessary to answer the question is concentrated in a single sentence or clustered tightly in a single paragraph.

The SQuAD [8] contains questions which are relatively simple because they are usually required no more than one sentence in a single paragraph to answer. SQuAD 2.0[9] introduces questions that are designed to be unanswerable. Bi-DAF (Seo et al., 2016) [18] and FastQA (Weissenborn et al., 2017) [19], which are popular for single-hop QA, the Query2Context and Context2Query modules in the Bi-DAF model are widely used in other QA models as core components. However, these models suffer dramatic accuracy declines in multi-hop QA task.

**Multi-hop Question Answering**

In general, two research directions have been explored to solve the multi-hop and multi-document QA task. The first direction is directed to apply the previous neural networks that are successful in single-hop QA tasks to multi-hop QA tasks. Zhong et al. (2019) [20] proposed a model combination coarse-grained reading and fine-grained reading. Query Focused Extractor model proposed by Nishida et al. (2019) [21] regards evidence extraction as a query-focused summarization task, and reformulates the query in each hop.

For complex questions, from the perspective of imitating human thinking, decomposing complex questions into simple subquestions is an effective method, Jiang and Bansel. [22] proposed a model for multi-hop QA, four atomic neural modules are designed, namely Find, Relocate, Compare, NoOp, where four neural modules were dynamically assembled to make multi-hop reasoning and support fact selection more interpretable. Concurrently to self-assembling modular networks, Min et al [15]. also addressed HotpotQA by decomposing its multi-hop questions into single-hop subquestions to achieve better performance and interpretability. However, their system approaches question decomposition by having a decomposer model trained via human labels.

A subset of approaches has introduced end-to-end frameworks explicitly designed to emulate the step-by-step reasoning process involved in multi-hop QA and MRC. The Kundu et al. [23] model constructs paths connecting questions and candidate answers and subsequently scores them through a neural architecture. Jiang et al. [24] also constructed a proposer used to proposes an answer from every root-to-leaf path in the reasoning tree, and the Evidence Assembler extracts a key sentence containing the proposed answer from every path and combines them to predict the final answer.

The other direction is based on graph neural networks (GNNs) [25]. GNNs have been shown to be successful on many NLP tasks, and recent papers have also examined complex QA using graph neural networks, including graph attention networks, graph recurrent networks, graph convolutional networks and their variants [26,27,28]. Cao et al. [29] proposed a bi-directional attention mechanism that was combined with an entity graph convolutional network to obtain the relation-aware representation of nodes for entity graphs. Qiu et al. [30] used a recurrent decoder that guides a dynamic exploration of Wikipedia links among passages to build an "evidence trail" leading to passage with the answer span.

The multilevel graph network can represent the information in the text in more detail, so the hierarchical graph network proposed by Fang et al., 2019[31] leverages a hierarchical graph representation of the background knowledge (i.e., question, paragraphs, sentences, and entities). Tu et al. [32] constructed a graph connecting sentences that are part of the same document, share noun-phrases and have named entities or noun phrases in common with the question, and then applied a GNN to the graph to rank the top entity as the answer. However, these approaches often fail to adequately capture the inherent structure of documents and discard masses of valuable structural information when transforming documents into graphs.

Documents unrelated to the complex query may affect the accuracy of the model. In the "select, answer, and explain" (SAE) model proposed by Tu et al. [33], BERT [34] acts as the encoder in the selection module. Then a sentence extractor is applied to the output of BERT to obtain the sequential output of each sentence with precalculated sentence start and end indices, to filter out answer-unrelated documents and thus reduce the amount of distraction information. The selected answer-related documents are then input to a model, which jointly predicts the answer and supporting sentences. Concurrently to the SAE model, Bhargav et al. [35] used a two-stage BERT-based architecture to first select the supporting sentence and then used the filtered supporting sentence to predict the answer. The upstream side of Jiang et al. [24] proposed model is the Document Explorer to iteratively address relevant documents.

**3 Task Definition**

| Input: | Query Q (text), $Q = \{q_1, q_2 \ldots q_J\}$ |
| --- | --- |
| | Context C (multiple texts), $C = \{x_1, x_2 \ldots x_T\}$ |
| Output: | Answer Type $A_T$ (label), |
| | Answer String $A_S$ (text), |
| | Supporting facts (multiple texts) |

Table 2: Symbol definition

As shown in Table 2, context C and query Q have T words and J words respectively, where C is regarded as one connected text. Q is regarded as a complex query. It is worth noting that when C is too long (e. g., over 2550 words) and should be truncated.

The multi-hop QA task is then defined as finding an answer string $A_S$, an answer type $A_T$ and support facts for a complex query. The answer type $A_T$ is selected from the answer candidates, such as 'yes/no/span'. The answer string $A_S$ is a short span in context, which is determined by predicting the positions of the start token and the end token when there are not enough answer candidates to answer Q. Supporting facts consist of one more than sentences in C and is required to answer Q.

# 4 Model

## 4.1 Overview

Our intuition is drawn from the human reasoning process for QA, and we propose a **C**oarse-**g**rain **De**composition **F**ine-**g**rain **in**teraction (CGDe-FGIn) model. The model mainly consists of context and question embedding layer, contextual embedding layer, coarse-grained decomposition layer, fine-grained interaction layer, modeling layer and output layer. We discuss each part separately in the next section. The overall model architecture is illustrated in Fig. 1.

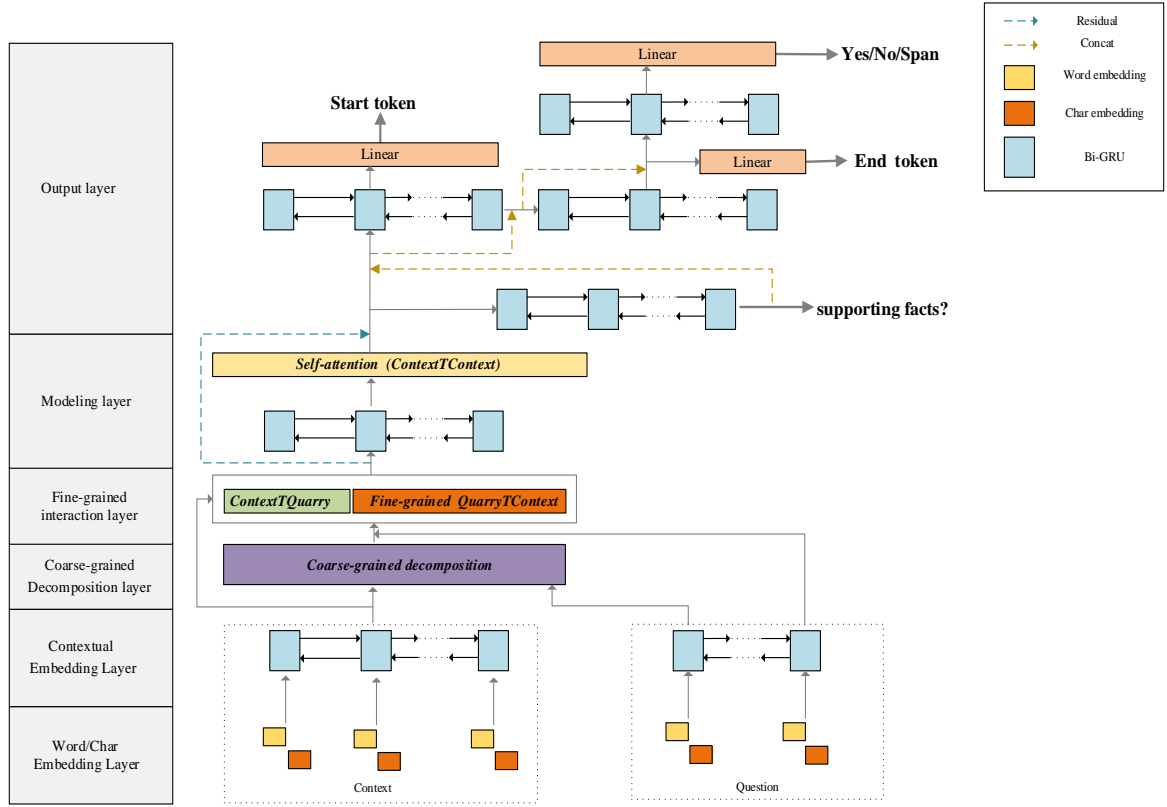

Figure 1: Overview of the CGDe-FGIn architecture

## 4.2 Context and Question Embedding Layer

We use a pre-trained word embedding model and a char embedding model to lay the foundation for CGDe-FGIn model. Let $\{x_1, x_2 \cdots x_T\}$ and $\{q_1, q_2 \cdots q_J\}$ represent the words in the input multi context paragraphs and complex query, respectively.

Following Yang et al. 2018[13] we use pre-trained word vectors in the form of GloVe (Pennington et al., 2014[36]) to obtain the fixed word embedding of each word, and we obtain the character level embedding of each word using convolutional neural networks (CNNs). The concatenation of the character and word embedding vectors is passed to a two-layer highway network (Srivastava et al., 2015[37]). The outputs of the highway network are two sequences of d dimensional vectors, or more conveniently, two matrices $X \in \mathbb{R}^{T \times d}$ for the context and $Q \in \mathbb{R}^{J \times d}$ for the query. where T and J are the numbers of words in multiple documents and queries respectively, and d is the dimension after fusion of the word embedding and character level embedding.

### 4.3 Contextual Embedding Layer

We use bi-directional recurrent neural networks with gated recurrent units (GRUs) (Cho et al., 2014[38]) to encode the contextual information present in the query and multiple context paragraphs separately. The outputs of the query and document encoders are $U \in \mathbb{R}^{J \times 2d}$ and $H \in \mathbb{R}^{T \times 2d}$, respectively. Here, 2d denotes the output dimension of the encoders. Note that each column vector of H and U has dimension 2d because of the concatenation of the outputs of the forward and backward GRUs, each with d-dimensional output.

### 4.4 Coarse-grained Decomposition Layer

Coarse-grained Decomposition layer is responsible for decomposing complex questions and generating new question high-dimensional vectors.

*Similarity matrix computatione*

First, a semantic similarity matrix is calculated for question(U) and multiple documents (H)as described by Yang et al [13]. Semantic similarity matrix $S \in \mathbb{R}^{T \times J}$, where Stj indicates the similarity between the t-th context word and j-th query word. The similarity matrix is computed by:

$$h = \text{linear}(\mathbf{H}) , \quad h \in \mathbb{R}^{T \times 1} \quad (1)$$
$$u = \text{permute}(\text{linear}(\mathbf{U})), \quad u \in \mathbb{R}^{1 \times J} \quad (2)$$
$$\alpha(\mathbf{H}, \mathbf{U}) = U^\top H , \quad \alpha(\mathbf{H}, \mathbf{U}) \in \mathbb{R}^{T \times J} \quad (3)$$
$$S_{tj} = [ h + u + \alpha(\mathbf{H}, \mathbf{U}) ], \quad S_{tj} \in \mathbb{R}^{T \times J} \quad (4)$$

where linear indicates a linear layer, permute represents vectors dimension transformation operations, ⊤ indicates matrix transpose.

Inspired by human hop-by-hop reasoning behavior, the meaning of complex questions decomposition is to make the high-dimensional vector distribution of entity nouns or pronouns more inclined to the intermediate answer to the question. For example, "The rapper whose debut album was titled "Thug Misses" has sold over how many records worldwide?", this relatively complex question can be decomposed into two subquestions, "Who is the rapper whose debut album was titled 'Thug Misses'?" and "How many records has that rapper sold worldwide?". Therefore, the answer to the first subquestion is crucial to answering the second question.

In answering complex questions, high-dimensional vectors for nouns such as "The Rapper" are expected to be more similar to intermediate answers required to answer the complex questions, such as "by America Rapper Khia." This is a disguised decomposition of a complex query.

To understand this point better, we transpose the $S_{tj}$ matrix to obtain $\tilde{S}_{jt}$. As shown in Fig. 2, the attention weight is computed by

$$a_{j:} = \text{softmax}(\tilde{S}_{j:}), a_{j:} \in \mathbb{R}^T \quad (5)$$

and query vector is computed by

$$\tilde{Q} = H^\top a, \quad \tilde{Q} \in \mathbb{R}^{J \times 2d} \quad (6)$$

Hence $\tilde{Q}$ is a J-by-2d matrix containing the attended context vectors for the entire query. To preserve the original information of the query, we fuse two vectors to obtain a new query representation. The representation is computed by

$$\bar{Q}=\beta(U;\tilde{Q}), \quad \bar{Q}\in\mathbb{R}^{J\times 2d} \tag{7}$$

$$\beta(U;\tilde{Q}) = W_{(S)}[\ U;\ \tilde{Q};\ U\circ\tilde{Q}] \tag{8}$$

where $W_{(S)}\in\mathbb{R}^{6d}$ is a trainable weight vector, ° represents elementwise multiplication, [;] represents vector concatenation across row, and implicit multiplication consists of matrix multiplication.

We obtain $\bar{Q}$, which is the integration of the original query and decomposed query, repeat the similarity matrix calculation, and then apply it to the subsequent model. The overall architecture is shown in Fig 3.

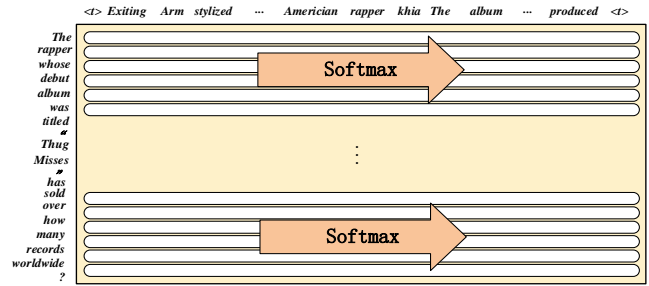

Figure 2: Similarity matrix softmax according to the query direction

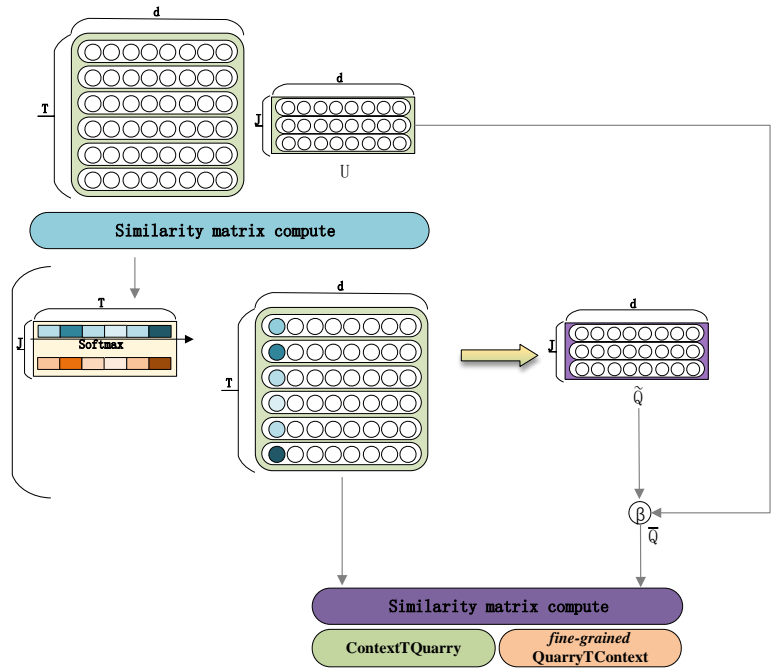

Figure3: Overview architecture of the Coarse-grained Decomposition layer

## 4.5 Fine-grained Interaction Layer

In the work of (Seo et al., 2017[18]), the *Query2Context* model component obtains the attention weights on the context words by $\mathbf{b} = \text{softmax}(\max_{\text{col}}(\mathbf{S})) \in \mathbb{R}^T$, where the maximum function ($\max_{\text{col}}$) is performed across the column. Then, the attended context vector is

$$\tilde{h} = \sum_t b_t H_{:t}, \quad \tilde{h} \in \mathbb{R}^{2d} \qquad (9)$$

This vector indicates the weighted sum of the most important words in the context with respect to the query. Here, $\tilde{h}$ is tiled T times across the column, thus giving $\tilde{H} \in \mathbb{R}^{T \times 2d}$, as shown in Fig 4.

The vanilla *Query2Context* module has two main deficiencies. First, the maximum function (max col) is performed across the column, and words that are consistent with the context in the question have a higher weight, such as the words "rapper" and "whose" in Fig 5. As a result, constituting middle answer words needed to answer complex questions, are easy to ignore, therefore, the original Query2Context model not perform well in supporting facts predicted task.

Second, since the size of the vector output of the vanilla Query2Context module is (batch size, 1, 2d), it needs to be repeated T times to obtain the vector of the same size as the input document, to meet the requirements of the vector size of subsequent model input. However, T times of repeated operations also result in the same high-dimensional vectors characteristics for each word in the contextual embedding of the context.

The output layer of the model classifies the word vector characteristics of each word in the context to evaluate the starting and ending positions of the answer; such output of the vanilla Query2Context is clearly not favorable to the subsequent model.

We introduce a method, as shown in Fig 6 to solve these problems. Instead of max pooling, softmax is used for each column of the attention matrix, and then the document vector is dotted with each column weight. The model obtains J vector matrices of size (T, 2d), where J is the number of words in the question, and where each matrix indicates the correlation between all words in the context and the corresponding word in the complex question. The similarity matrix $\bar{S}$ between the contextual embeddings of the context (H) and the new query ($\bar{Q}$) is computed by:

$$\bar{q} = \text{permute}(\text{linear}(\bar{Q})), \quad \bar{q} \in \mathbb{R}^{1 \times J} \qquad (10)$$

$$\bar{S}_{tj} = [h + \bar{q} + \alpha(\mathbf{H}, \bar{Q})], \quad \bar{S}_{tj} \in \mathbb{R}^{T \times J} \qquad (11)$$

the attention weight $\bar{a}$ is computed by:

$$\bar{a}_{:j} = \text{softmax}(\bar{S}_{:j}), \quad \bar{a}_{:j} \in \mathbb{R}^T \qquad (12)$$

The fine-grained Query2Context representation $\bar{U}$ is computed by:

$$\bar{U} = \sum_j \bar{a}_{:j} \circ H, \quad \bar{U} \in \mathbb{R}^{T \times 2d} \qquad (13)$$

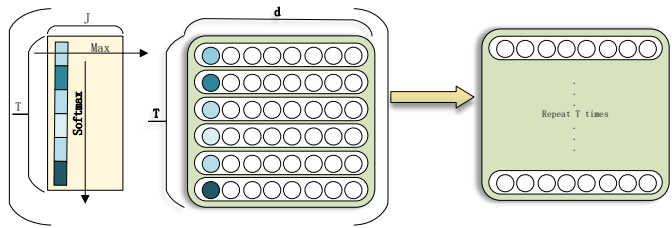

Figure 4: Vanilla Query2Context

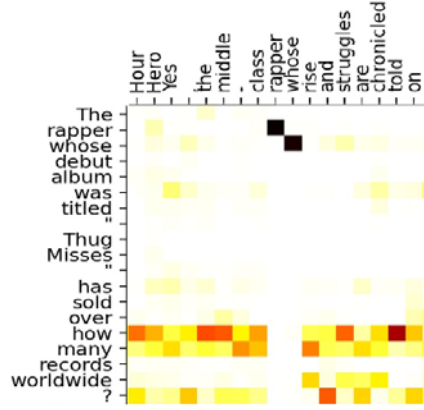

Figure 5: Heatmap of the semantic similarity matrix

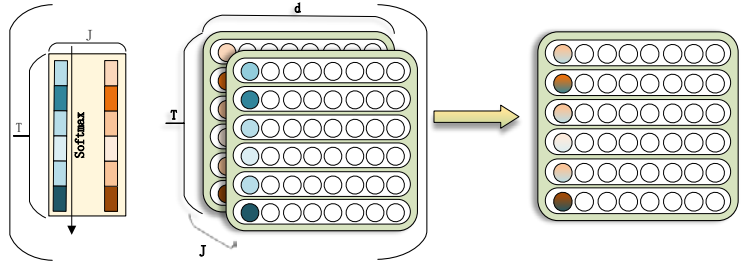

Figure 6: Fine-grained interaction

The Context2Query attention signifies which query words are most relevant to each context word, which is computed by:

$$\bar{a}_t = \text{softmax}(\bar{S}_{t:}), \ \bar{a}_t \in \mathbb{R}^J \qquad (14)$$

$$\tilde{U}_{:t} = \sum_j \bar{a}_{tj} U_{:j} \qquad (15)$$

Finally, the contextual embeddings and the feature vectors computed by the fine-grained interaction layer are combined together to yield G:

$$G_{:t} = \tilde{\beta}(\mathbf{H}_{:t}, \tilde{U}_{:t}, \bar{U}_{:t}) \qquad (16)$$

$$\tilde{\beta}(h, \tilde{u}, \bar{u}) = [\,h;\ \tilde{u};\ h \circ \bar{u};\ \bar{u} \circ \tilde{u}\,] \qquad (17)$$

**4.6 Modeling Layer**

The output G of the fine-grained QueryTcontext layer is taken as input to the modeling layer, which encodes the query-aware representations of context words. We use one layers of the bi-directional GRU to capture the interaction among the context words conditioned on the query. Since multiple documents contain thousands of words, the long-distance dependency problem is obvious, so a self-attention module is added to alleviate this problem. Similar to the baseline model, we use the original Bi-DAF function to implement self-attention, in which the input is changed from (query, context) to (context, context).

**4.7 Prediction Layer**

We follow the same structure of prediction layers as (Yang et al., 2018[13]). To solve the degradation problem of the deep neural network, residual connections are made between the output of the fine-grained

QueryTcontext layer and the output of the modeling layer, which is the input to the prediction layer. Within the prediction layer, four isomorphic Bi-GRUs are stacked layer by layer, and we adopt a cascade structure to solve the output dependency problem and avoid information loss.

The prediction layer has four output dimensions: 1. supporting sentences, 2. the start position of the answer, 3. the end position of the answer, and 4. the answer type. Depending on the type, different architectures are chosen. In this work, we investigate all of the above prediction types.

We define the training loss (to be minimized) as:

$$\mathcal{L}_{type} = \text{CE}_{sum}(\hat{y}_{type}, y_{type}) \quad \mathcal{L}_{sup} = \text{CE}_{average}(\hat{y}_{sup}, y_{sup})$$
$$\mathcal{L}_{start} = \text{CE}_{sum}(\hat{y}_{start}, y_{start}) \quad \mathcal{L}_{end} = \text{CE}_{sum}(\hat{y}_{end}, y_{end})$$

We jointly optimize these four cross entropy losses, and each loss term is weighted by a coefficient

$$\mathcal{L} = \lambda_a (\mathcal{L}_{type} + \mathcal{L}_{start} + \mathcal{L}_{end}) + \lambda_s \mathcal{L}_{sup}$$

## 5 Experiments

### 5.1 Datasets

Two publicly available QA datasets are employed to test the performance of the proposed model.

**HotpotQA** is a recently introduced multi-hop RC dataset encompassing Wikipedia articles, and there are two types of questions—bridge and comparison. We evaluate our model on development sets in the distractor setting, following prior work. For the full wiki setting where all Wikipedia articles are given as input, we consider the bottleneck to be about information retrieval, thus we do not include the full wiki setting in our experiments.

**SQuAD 1.1** contains 100K crowdsourced questions and answers paired with short Wikipedia passages. The typical length of the paragraphs is approximately 250 and the question is 10 tokens although there are exceptionally long cases. The SQuAD dataset is mainly used to verify the validity and universality of the model components we propose, namely coarse-grained decomposition strategy and fine-grained interaction strategy.

For both HotpotQA and SQuAD 1.1, only the training and validation data are publicly available, while the test data are hidden. For further analysis, we report only the performance on the validation set, as we do not want to probe the unseen test set by frequent submissions. According to the observations from our experiments and previous works, the validation score is well correlated with the test score.

### 5.2 Implementation Details

We keep the baseline (Bi-DAF) parameter settings on the two data sets to prove that our model components and model architecture have absolute performance advantages over the baseline.

For the HotpotQA dataset, we use the standard 300-dimensional pre-trained GloVe (trained from 840B web crawled data points) as word embeddings. The dimensions of hidden states in BiGRU are set as $d = 80$. Using the Adam optimizer, with a minibatch size of 32 and an initial learning rate of 0.01, an early stopping strategy is adopted, with patience=1, and $\lambda_a = 0.5$, $\lambda_s = 2.0$. The training process takes approximately 8 hours on two 2080 ti GPUs.

For the SQuAD dataset, we also use the standard 300-dimensional pre-trained GloVe as word embeddings. The hidden state size $d = 100$, using the the AdaDelta optimizer, with a minibatch size of 32 and an initial learning rate of 0.5. A dropout (Srivastava et al., 2014[39]) rate of 0.2 is used for the CNN and LSTM layers, and the linear transformation before the softmax for the answers. During training, the moving averages of all weights of the model are maintained with an exponential decay rate of 0.999. The training process takes approximately 6 hours on a single 2080 ti GPU.

## 5.3 Main Results

**Model Comparison**

We compare the results with those of two types of baseline model. One is the model with Bi-DAF as the core component. Questions and documents are not processed by off-the-shelf language tools, but only contextual embedding is performed. This type of models is dedicated mainly to the feature interaction between questions and documents. The advantages of these models are fewer model parameters, short training time, and low GPU computing power requirements.

The other is the reasoning model based on a graph neural network. This type of model usually uses a language model or tool for named entity recognition to construct an entity graph, and then a graph convolutional neural network is used to update the node representation on the entity graph. The output layer uses a classifier to determine whether the entity is the correct answer. The effect of this type of model is generally higher than that of the first type of model, and it has relatively high interpretability. However, the premise assumes that the answers to complex questions are entities, and they are all in the constructed graph network. These models also need to use tools to extract entities from multiple documents, which increases the training time and heightens GPU requirements.

| Model | Answer | | Sup Fact | | Joint | |
|---|---|---|---|---|---|---|
| | EM | F1 | EM | F1 | EM | F1 |
| Baseline | 45.60 | 59.02 | 20.32 | 64.49 | 10.83 | 40.16 |
| NMN | 49.58 | 62.71 | - | - | - | - |
| KGNN | 50.81 | 65.75 | 38.74 | 76.69 | 22.40 | 52.82 |
| Our Model | **50.89±0.13** | 65.41±0.18 | **39.47±0.46** | **79.83±0.14** | **23.08±0.39** | **54.51±0.29** |

Table 3: The performance of our CGDe-FGIn model and competing approaches by Yang et al., and Ye et al., Jiang et al. on the HotpotQA dataset.

The performance of mul-hop QA on HotpotQA is evaluated by using the exact match (EM) and F1 as two evaluation metrics. To assess the explainability of the models, the datasets further introduce two sets of database metrics involving the supporting facts. The first set focuses on evaluating the supporting facts directly, namely EM and F1 on the set of supporting fact sentences compared to the gold set. The second set features joint metrics that combine the evaluation of answer spans and supporting facts. All metrics are evaluated example-by-example, and then averaged over examples in the evaluation set.

We compare our approach with several previously published models, and present our results in Table 3. All experiments are performed for each of our models, and the table shows the mean and standard deviation. As shown in the table, all the results of our proposed model are superior to those of the baseline model in the case that the model parameters are not increased substantially.

## 4.4 Ablations Studies

In this paper, we design two strategies for multi-hop Question Answering. To study the contributions of these two strategies to the performance of our model, we conduct an ablation experiment by removing coarse-grained decomposition strategy or fine-grained interaction strategy on the SQuAD1.1 and HotpotQA datasets.

| Model | Answer | | Sup Fact | | Joint | |
|---|---|---|---|---|---|---|
| | EM | F1 | EM | F1 | EM | F1 |
| Baseline | 45.60 | 59.02 | 20.32 | 64.49 | 10.83 | 40.16 |
| Our Model | **50.89±0.13** | **65.41±0.18** | 39.47±0.46 | 79.83±0.14 | **23.08±0.39** | **54.51±0.29** |
| CGDe | 50.55±0.22 | 65.27±0.11 | 38.79±0.28 | 79.26±0.14 | 22.48±0.29 | 53.87±0.15 |
| FGIn | 50.07±0.64 | 64.61±0.33 | **40.55±0.42** | **80.55±0.18** | 22.94±0.21 | 54.12±0.33 |

Table 4: Ablation results on the HotpotQA dev set.

| Model | EM | F1 |
|---|---|---|
| Baseline Model | 64.56 | 75.51 |
| FGIn | 66.32 | 76.93 |
| CGDe | 65.25 | 75.96 |
| CGDe / FGQTC | **66.44** | **77.06** |

Table 5: Ablation results on the SQuAD dev set.

As shown in Tables 4 and 5, removing either the CGDe or the FGIn strategy reduces the effectiveness of the model, which demonstrates that both strategies contribute to our model. Moreover, using either strategy individually enables our model to achieve better results than the baseline model.

**Analysis and Visualization**

In this section, we conduct a series of visual analyses with different settings using our approach.

*Coarse-grained decomposition*

The coarse-grain decomposition module uses the similarity matrix of the query and the document to be multiplied by the document representation to obtain a new query representation (J, 2d). After merging with the original query representation, the new query representation should have higher semantic similarity with the document's corresponding words, for example, the phrase " The rapper " and the word "Khia" in the complex question "The rapper whose debut album was titled 'Thug Misses' has sold over how many records worldwide?".

| |
|---|
| **Q1** Who is the rapper whose debut album was titled 'Thug Misses'? |
| Support fact one：Thug Misses is the debut album by American rapper Khia. |
| **Q2** How many records has that rapper sold worldwide? |
| Support fact two：To date Khia has collectively sold over 2 million records worldwide. |

Table 6: Subquestion and Supporting facts

As the subquestion and supporting facts shown in Table 6, we hope that the phrase "The rapper" and the word "Khia" have more similar expressions, so that complex queries become simple one-hop queries: " The rapper (Khia) whose debut album was titled 'Thug Misses'has sold over how many records worldwide ".

To confirm our idea, we use the baseline trained model and our model to process the validation set and generate the heat map of the attention matrix (the darker the color in the figure, the higher is the similarity weight), respectively.

In the baseline model's heat map, the attention weights of the phrase "The rapper" and the word "Khia" are not high, it is worth noting that this is caused by the similarity of the parts of speech between the two phrases, the part of speech of "rapper" is a noun, while the part of speech of "Khia" is a person's name, resulting in a slightly higher correlation between the two phrases. Different from the baseline model, the heat map of our model shows that the semantic similarity of the phrase "The rapper" and the word "Khia" is significantly higher than that of other surrounding words. This shows that the new question contains the subanswers that appear in the text to a certain extent, so that the multi-hop query is decomposed into a simple single-hop query.

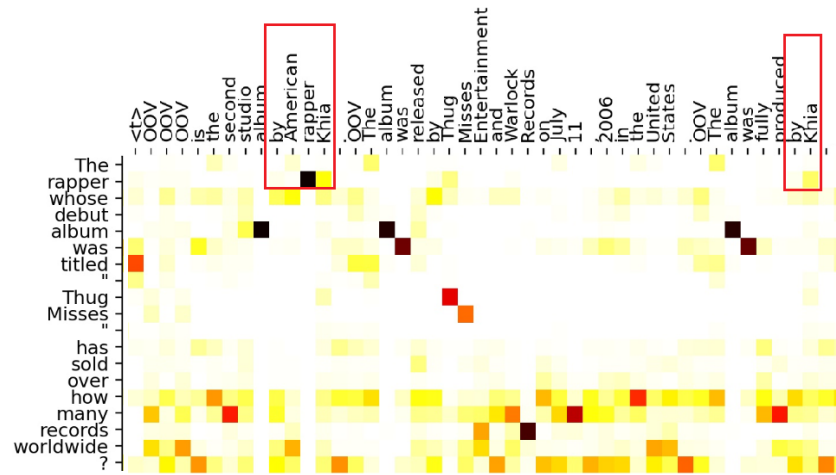

Figure 7: Attention heat map of the baseline model

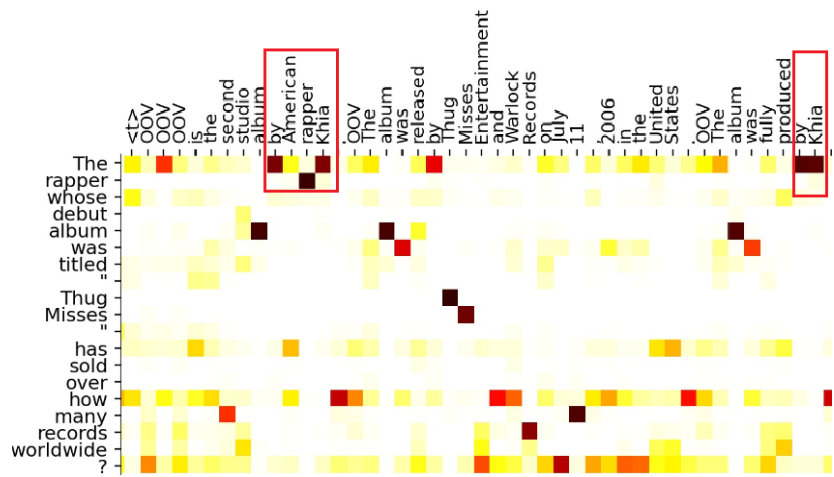

Figure 8: Attention heat map of our model

In the ablation study, it can be easily found that the coarse-grained decomposition module improves the EM and F1 of the answer in evaluation metrics; compared with the fine-grained interaction model, Sup Facts's EM and F1 have lower improvement. This shows that the model's ability to predict support facts is limited, because the new question generated contains the intermediate answer required for the first subquestion, so the support context that answers the first question may not be predicted as a supporting fact.

*Fine-grained interaction*

As shown in Table 4, the fine-grained interaction strategy performs well on the supporting facts task, which further proves that the strategy can model more appropriate semantic features represented by a high-dimensional vector for individual words in multiple documents. To make this more intuitive, we visually present the instances in HotpotQA datasets. According to the previous section, the complex query in Table 1 requires two supporting fact sentences, "Thug Misses is the debut album by American rapper Khia." and "To date Khia has collectively sold over 2 million records worldwide."

Fig. 9, (a) and (b) subgraph show heatmaps of the semantic similarity matrix of the baseline model (Bi-DAF), showing the part of the complex query corresponding to the supporting fact sentence. Similarly, subfigures (c) and (d) show the same part of our model with the fine-grained interaction strategy. Compared with the baseline model, the supporting fact sentences in our model have a higher weight in multiple documents.

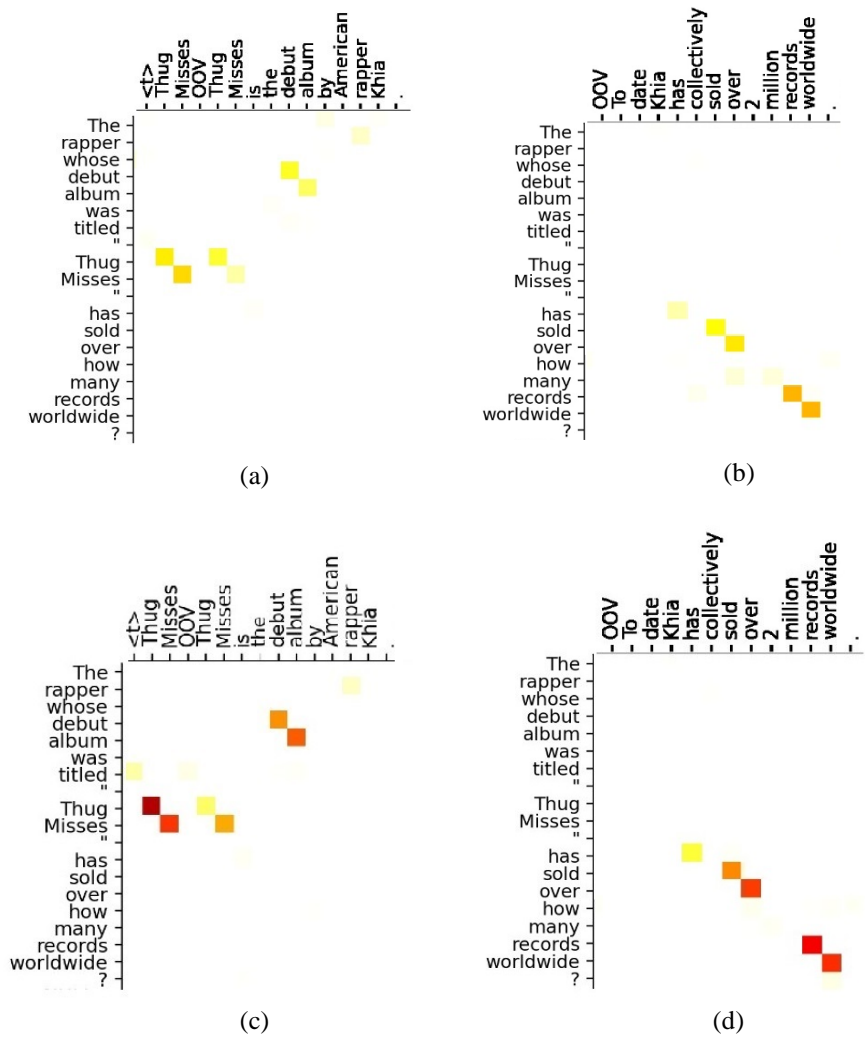

Figure 9: Attention heat map with the FGIn

## 6 Conclusion and Future Work

In this paper, we propose a mutli-hop question answering model, that contains a coarse-grained decomposition strategy to divide a complex query into multiple single-hop simple queries and a fine-grained interaction strategy to better represent each word in the document and help the model find the sentences needed to answer the question. In the experiments, we show that our models significantly and consistently outperform the baseline model.

In the future, we think that the following issues would be worth studying:

In Fine-grained interaction layer, assigning different weights to J context representations corresponding to each word in a complex query instead of adding them together can further improve our model. We plan to explore how to measure the importance of each word in the query at different stages of reasoning.

**Acknowledgments**

This research was funded by the Fundamental Research Funds for the Central Universities (Grant number 2020YJS012).

**References**

[1] Danqi Chen, Adam Fisch, Jason Weston, and Antoine Bordes. 2017. Reading wikipedia to answer open domain questions. In ACL.
[2] Xiong C, Zhong V, Socher R. Dynamic coattention networks for question answering[J]. arXiv preprint arXiv:1611.01604, 2016.
[3] Cui Y, Chen Z, Wei S, et al. Attention-over-Attention Neural Networks for Reading Comprehension[C]//Proceedings of the 55th Annual Meeting of the Association for Computational Linguistics (Volume 1: Long Papers). 2017: 593-602.
[4] Huang H Y, Zhu C, Shen Y, et al. Fusionnet: Fusing via fully-aware attention with application to machine comprehension[J]. arXiv preprint arXiv:1711.07341, 2017.
[5] Huang M, Zhu X, Gao J. Challenges in building intelligent open-domain dialog systems[J]. ACM Transactions on Information Systems (TOIS), 2020, 38(3): 1-32.
[6] Chen H, Liu X, Yin D, et al. A survey on dialogue systems: Recent advances and new frontiers[J]. Acm Sigkdd Explorations Newsletter, 2017, 19(2): 25-35.
[7] Liu N, Shen B. ReMemNN: A novel memory neural network for powerful interaction in aspect-based sentiment analysis[J]. Neurocomputing, 2020.
[8] Rajpurkar, P.; Zhang, J.; Lopyrev, K.; and Liang, P. 2016. Squad: 100,000+ questions for machine comprehension of text. In Proceedings of the 2016 Conference on Empirical Methods in Natural Language Processing, 2383–2392.
[9] Rajpurkar P, Jia R, Liang P. Know What You Don't Know: Unanswerable Questions for SQuAD[C]//Proceedings of the 56th Annual Meeting of the Association for Computational Linguistics (Volume 2: Short Papers). 2018: 784-789.
[10] Kočiský, T.; Schwarz, J.; Blunsom, P.; Dyer, C.; Hermann, K. M.; Melis, G.; and Grefenstette, E. 2018. The narrativeqa reading comprehension challenge. Transactions of the Association of Computational Linguistics 6:317–328.
[11] Reddy, S.; Chen, D.; and Manning, C. D. 2019. Coqa: A conversational question answering challenge. Transactions of the Association for Computational Linguistics 7:249–266.
[12] Lin X V, Socher R, Xiong C. Multi-hop knowledge graph reasoning with reward shaping[J]. arXiv preprint arXiv:1808.10568, 2018.
[13] Zhilin Yang, Peng Qi, Saizheng Zhang, Yoshua Bengio, William Cohen, Ruslan Salakhutdinov, and

Christopher D Manning. 2018. Hotpotqa: A dataset for diverse, explainable multi-hop question answering. In Proceedings of the 2018 Conference on Empirical Methods in Natural Language Processing, pages 2369–2380.

[14] Welbl J, Stenetorp P, Riedel S. Constructing datasets for multi-hop reading comprehension across documents[J]. Transactions of the Association for Computational Linguistics, 2018, 6: 287-302.

[15] Sewon Min, Victor Zhong, Luke Zettlemoyer, and Hannaneh Hajishirzi. 2019b. Multi-hop reading comprehension through question decomposition and rescoring. In ACL.

[16] Qi P, Lin X, Mehr L, et al. Answering complex open-domain questions through iterative query generation[J]. arXiv preprint arXiv:1910.07000, 2019.

[17] Jiang Y, Bansal M. Avoiding reasoning shortcuts: Adversarial evaluation, training, and model development for multi-hop QA[J]. arXiv preprint arXiv:1906.07132, 2019.

[18] Minjoon Seo, Aniruddha Kembhavi, Ali Farhadi, and Hannaneh Hajishirzi. 2017. Bidirectional attention flow for machine comprehension. In Proceedings of the International Conference on Learning Representations.

[19] Weissenborn D, Wiese G, Seiffe L. Fastqa: A simple and efficient neural architecture for question answering[J]. arXiv preprint arXiv:1703.04816, 2017.

[20] Victor Zhong, Caiming Xiong, Nitish Shirish Keskar, and Richard Socher. 2019. Coarse-grain fine-grain coattention network for multi-evidence question answering. In ICLR.

[21] Nishida K, Nishida K, Nagata M, et al. Answering while Summarizing: Multi-task Learning for Multi-hop QA with Evidence Extraction[C]//Proceedings of the 57th Annual Meeting of the Association for Computational Linguistics. 2019: 2335-2345.

[22] Jiang Y, Bansal M. Self-Assembling Modular Networks for Interpretable Multi-Hop Reasoning[C]//Proceedings of the 2019 Conference on Empirical Methods in Natural Language Processing and the 9th International Joint Conference on Natural Language Processing (EMNLP-IJCNLP). 2019: 4464-4474.

[23] Kundu S, Khot T, Sabharwal A, et al. Exploiting Explicit Paths for Multi-hop Reading Comprehension[C]//Proceedings of the 57th Annual Meeting of the Association for Computational Linguistics. 2019: 2737-2747.

[24] Jiang Y, Joshi N, Chen Y C, et al. Explore, Propose, and Assemble: An Interpretable Model for Multi-Hop Reading Comprehension[C]//Proceedings of the 57th Annual Meeting of the Association for Computational Linguistics. 2019: 2714-2725.

[25] Xu K, Hu W, Leskovec J, et al. How powerful are graph neural networks?[J]. arXiv preprint arXiv:1810.00826, 2018.

[26] Veličković P, Cucurull G, Casanova A, et al. Graph attention networks[J]. arXiv preprint arXiv:1710.10903, 2017.

[27] Hajiramezanali E, Hasanzadeh A, Narayanan K, et al. Variational graph recurrent neural networks[C]//Advances in neural information processing systems. 2019: 10701-10711.

[28] Kipf T N, Welling M. Semi-supervised classification with graph convolutional networks[J]. arXiv preprint arXiv:1609.02907, 2016.

[29] Cao Y, Fang M, Tao D. BAG: Bi-directional Attention Entity Graph Convolutional Network for Multi-hop Reasoning Question Answering[C]//Proceedings of the 2019 Conference of the North American Chapter of the Association for Computational Linguistics: Human Language Technologies, Volume 1 (Long and Short Papers). 2019: 357-362.

[30] Qiu L, Xiao Y, Qu Y, et al. Dynamically fused graph network for multi-hop reasoning[C]//Proceedings of the 57th Annual Meeting of the Association for Computational Linguistics. 2019: 6140-6150.

[31] Y. Fang, S. Sun, Z. Gan, R. Pillai, S. Wang, and J. Liu. Hierarchical graph network for multi-hop question answering. arXiv preprint arXiv:1911.03631, 2019.

[32] Tu M, Wang G, Huang J, et al. Multi-hop Reading Comprehension across Multiple Documents by Reasoning over Heterogeneous Graphs[C]//Proceedings of the 57th Annual Meeting of the Association for Computational Linguistics. 2019: 2704-2713.

[33] Tu M, Huang K, Wang G, et al. Select, Answer and Explain: Interpretable Multi-Hop Reading Comprehension over Multiple Documents[C]//AAAI. 2020: 9073-9080.

[34] Devlin J, Chang M W, Lee K, et al. Bert: Pre-training of deep bidirectional transformers for language understanding[J]. arXiv preprint arXiv:1810.04805, 2018.

[35] Bhargav G P S, Glass M, Garg D, et al. Translucent Answer Predictions in Multi-Hop Reading Comprehension[C]//Proceedings of the AAAI Conference on Artificial Intelligence. 2020, 34(05): 7700-7707.

[36] Pennington J, Socher R, Manning C D. Glove: Global vectors for word representation[C]//Proceedings of the 2014 conference on empirical methods in natural language processing (EMNLP). 2014: 1532-1543.

[37] Srivastava R K, Greff K, Schmidhuber J. Highway networks[J]. arXiv preprint arXiv:1505.00387, 2015.

[38] Chung J, Gulcehre C, Cho K H, et al. Empirical evaluation of gated recurrent neural networks on sequence modeling[J]. arXiv preprint arXiv:1412.3555, 2014.

[39] Nitish Srivastava, Geoffrey E. Hinton, Alex Krizhevsky, Ilya Sutskever, and Ruslan Salakhutdinov. Dropout: a simple way to prevent neural networks from overfitting. JMLR, 2014.